\documentclass[letterpaper, 10 pt, conference]{ieeeconf}  % Comment this line out if you need a4paper

\IEEEoverridecommandlockouts                              % This command is only needed if 
\usepackage{times}
\usepackage{amsmath,amssymb,amsfonts, cases}
\usepackage{mathtools}
\usepackage{soul}
\usepackage{array}
\usepackage{graphicx}
\usepackage{caption}
\usepackage{subcaption}
\usepackage[short]{optidef}
\usepackage[natbib=true]{biblatex}
\AtEveryBibitem{%
  \clearfield{issn} % Remove issn

  \ifentrytype{online}{}{% Remove url except for @online
    \clearfield{url}
  }
}
\usepackage{cuted}
\usepackage{lipsum}
\usepackage{multicol}
\usepackage{booktabs}
\usepackage{pifont}
\newcommand{\cmark}{\ding{51}}%
\newcommand{\xmark}{\ding{55}}%
\usepackage[bookmarks=true,hidelinks]{hyperref}
\usepackage[usenames, dvipsnames]{color}

\newcommand{\diff}[1]{\textcolor{blue}{#1}}

\usepackage{multirow}
\usepackage[utf8]{inputenc}
\usepackage{pgfplots} 
\usepackage{pgfgantt}
\usepackage{pdflscape}
\pgfplotsset{compat=newest} 
\pgfplotsset{plot coordinates/math parser=false}
\pgfplotsset{yticklabel style={
        /pgf/number format/fixed
},scaled y ticks=false}
\newlength\fheight
\newlength\fwidth
\title{\LARGE \bf
SLoMo: A General System for Legged Robot Motion Imitation from Casual Videos
\vspace*{-5mm}
}
\author{John Z. Zhang$^{1}$, Shuo Yang$^{2}$, Gengshan Yang$^{1}$, Arun L. Bishop$^{1}$, Swaminathan Gurumurthy$^{1}$,\\ Deva Ramanan$^{1}$, and Zachary Manchester$^{1}$
\thanks{$^{1}$ Robotics Institute, $^{2}$ Mechanical Engineering, CMU, Pittsburgh, PA}
\thanks{Email: johnzhang@cmu.edu}
}
\begin{document}

\maketitle
\begin{strip}
  \centering
  \vspace*{-20mm}
  \includegraphics[width=\linewidth]{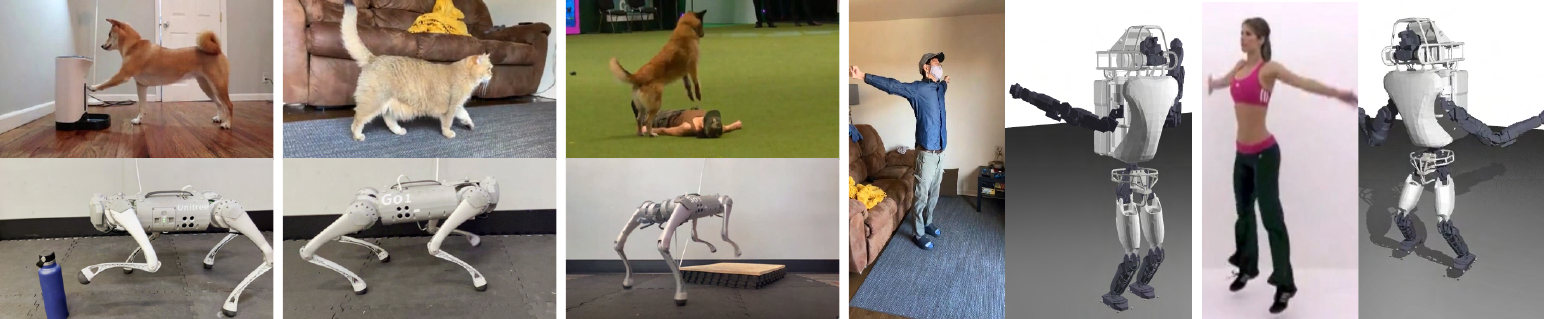}
  \captionof{figure}{A collection of video-to-robot motion-transfer demonstrations on the Unitree Go1 quadrupedal robot on hardware (left three) and the Atlas humanoid robot in simulation (right two). From left to right: a dog reaching for a water feeder with one of its front feet, a house cat pacing, a trained dog performing a Cardiopulmonary Resuscitation (CPR) exercise on its human partner, a human stretching his body and limbs, and a human demonstrating a jumping-jack exercise.}
  \label{fig:teaser-demo}
  \vspace*{-5mm}
\end{strip}
% \twocolumn[  \maketitle    \begin{figure*}[h]
%     \centering
%     \includegraphics[width=\textwidth]{figures/intro_vid_robot_comp.pdf}
%     \caption{Caption for the figure.}
%   \end{figure*}
% ]
% \onecolumn
% \twocolumn[{%
% \renewcommand\twocolumn[1][]{#1}%
% \maketitle
% \vspace*{-5mm}
% \begin{center}
%     \centering
%     \captionsetup{type=figure}
%     \includegraphics[width=1.0\textwidth]{figures/intro_vid_robot_comp.pdf}
%     \captionof{figure}{A collection of video-to-robot motion transfer demonstrations on the Unitree Go1 quadrupedal robot on hardware (left three) and Atlas humanoid robot in simulation (right). From left to right: a dog reaching for a water feeder with one of its front feet, a house cat pacing gracefully across the living room floor, a trained dog performing a Cardiopulmonary Resuscitation (CPR) exercise on its human partner during a competition routine, and a human stretching his body and limbs.}
%     \label{fig:teaser-demo}
% \end{center}%
% }]

\thispagestyle{plain}
\pagestyle{plain}

%%%%%%%%%%%%%%%%%%%%%%%%%%%%%%%%%%%%%%%%%%%%%%%%%%%%%%%%%%%%%%%%%%%%%%%%%%%%%%%%
% \begin{abstract}

% This electronic document is a ÒliveÓ template. The various components of your paper [title, text, heads, etc.] are already defined on the style sheet, as illustrated by the portions given in this document.

% \end{abstract}

\begin{figure*}[h]
    \centering
    \includegraphics[width=0.8\textwidth]{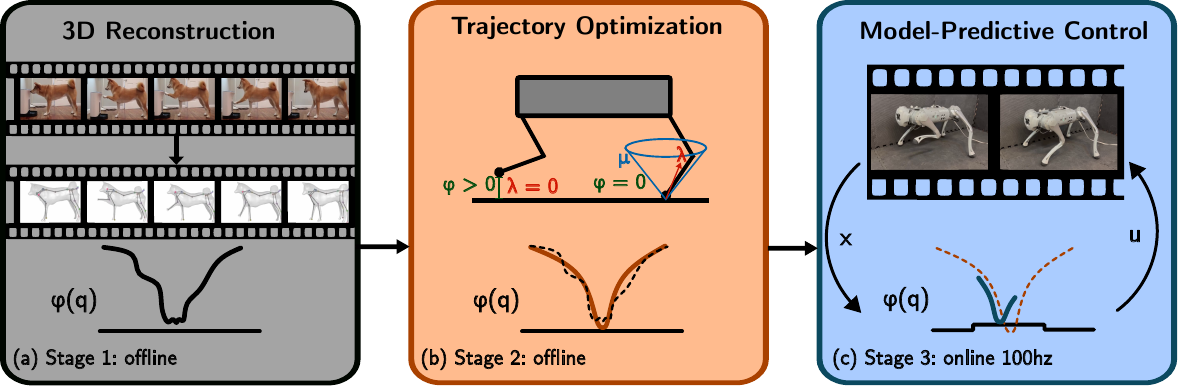}
    \caption{The SLoMo algorithm: (a) Stage 1: Process visual RGB inputs and generate body and foot trajectories in $\mathbb{R}^3$. This reconstruction considers physics but does not produce a trajectory that is dynamically feasible on the target robot. (b) Stage 2: Solve for open-loop robot state, control, and contact-force reference trajectories that imitate the 3D reconstruction while obeying robot dynamics and contact constraints. (c) Stage 3: Track the reference trajectory on robot hardware while managing model and contact-timing mismatch and disturbances.}
    \label{fig:method_overview}
    \vspace*{-5mm}
\end{figure*}
% \vspace*{-10mm}
\begin{abstract}
We present SLoMo: a first-of-its-kind framework for transferring skilled motions from casually captured ``in-the-wild" video footage of humans and animals to legged robots. SLoMo works in three stages: 1) synthesize a physically plausible reconstructed key-point trajectory from monocular videos; 2) optimize a dynamically feasible reference trajectory for the robot offline that includes body and foot motion, as well as a contact sequence that closely tracks the key points; 3) track the reference trajectory online using a general-purpose model-predictive controller on robot hardware. Traditional motion imitation for legged motor skills often requires expert animators, collaborative demonstrations, and/or expensive motion-capture equipment, all of which limit scalability. Instead, SLoMo only relies on easy-to-obtain videos, readily available in online repositories like YouTube. It converts videos into motion primitives that can be executed reliably by real-world robots. We demonstrate our approach by transferring the motions of cats, dogs, and humans to example robots including a quadruped (on hardware) and a humanoid (in simulation). Videos are available at \href{https://slomo-www.github.io/website}{https://slomo-www.github.io/website}.
\diff{% To the best knowledge of the authors, this is the first attempt at a general-purpose motion transfer framework that imitates animal and human motions on legged robots directly from casual videos without artificial markers or labels. 
}
\end{abstract}
% \vspace{-2mm}

%%%%%%%%%%%%%%%%%%%%%%%%%%%%%%%%%%%%%%%%%%%%%%%%%%%%%%%%%%%%%%%%%%%%%%%%%%%%%%%%
\vspace{-1mm}
\section{Introduction}
\vspace{-1mm}
One of the grand challenges in robotics is to enable human- and animal-level agility for legged robots by directly imitating their natural counterparts. This motion-imitation procedure typically involves three steps: extracting motion primitives from videos or image sequences, processing motion primitives to ensure they are within the physical limits of the robot and, finally, executing those movements on robot hardware.  Prior work on motion imitation relies on marker-based, multi-camera motion-capture (MoCap) systems, limiting the diversity of captured movements. An end-to-end motion-transfer solution that takes in raw video data and executes novel behaviors on real-world robots has, so far, remained elusive.        

In this paper, we leverage recent advancements in neural rendering and reconstruction, trajectory optimization, and model-predictive control to build, to the best of the authors' knowledge, the first successful general framework for transferring human and animal motion skills captured by a single, moving camera to robot hardware. The framework, illustrated in Fig. \ref{fig:method_overview}, contains three modules: 1) a reconstruction pipeline that produces physically plausible 3D key-point trajectories from casual video footage;
2) a trajectory optimizer that solves for dynamically feasible robot state, control, and contact-force reference trajectories that closely mimic the key-point trajectories while respecting the physical limitations of the robot; and 3) a model-predictive controller (MPC) that runs at real-time rates on robot hardware to track reference trajectories. 
An important feature of our approach is explicit reasoning about contact interactions between the robot and environment at every stage of the framework, ensuring that offline reference trajectories can be safely executed on hardware and that the online MPC is robust to contact-timing and model mismatch. Additionally, model-based trajectory generation and control methods allow us to maintain explainability in each stage of the pipeline --- something difficult to achieve with current black-box reinforcement learning (RL) approaches. Finally, our work also differs from prior works in its generality across robot morphology; our 3D reconstruction pipeline, trajectory optimizer, and MPC are all robot agnostic --- enabling motion transfer from humans to humanoid robots and from animals to quadrupedal robots within a single, unified framework. 

Our specific contributions are:
\begin{itemize}
    \item A general-purpose motion-transfer framework, SLoMo, for enabling legged robots to mimic human and animal motions from casual videos 
    \item A novel offline reference-trajectory and contact-sequence generation technique that ensures physical feasibility
    \item End-to-end experimental demonstrations transferring animal behaviors to a quadruped robot on hardware and human motions to a humanoid robot in simulation
\end{itemize}

This paper is organized as follows: We review related literature in Section \ref{related_work}. Our methodology is introduced in Section \ref{motion_transfer}. Results of simulation and hardware experiments are reported in Section \ref{results}. Finally, we summarize our conclusions and discuss directions for future research in Section \ref{conclusions}.

\section{Background and Related Work}\label{related_work}
In this section, we review related literature on human and animal motion capture, trajectory optimization through contact, and online control for legged robots. 

\subsection{Human and Animal Motion Capture}\label{sec:motion-capture}
To study human and animal body movements, MoCap systems have been widely adopted in the movie industry and research labs~\cite{
% zhang2018mode, 
kang2022animal, luo2022artemis, cmumocap
% , park2006capturing
}. An optical MoCap system typically consists of multiple high-resolution cameras whose positions and orientations are precisely measured through a sophisticated calibration process. Multiple cameras can observe and triangulate the positions of markers, which can be mounted on human or animal subjects. Although some MoCap datasets have been made publicly available \cite{cmumocap}, obtaining novel motion sequences remains a challenge. Despite these shortcomings, most existing works in motion imitation \cite{bin_peng_learning_2020, kang_animal_2021
% , kang_animal_nodate
} rely on MoCap data. Marker-based MoCap systems have inherently limited capture volume and are very sensitive to camera configuration changes, making outdoor usage extremely difficult and unreliable. MoCap systems also typically cost tens to hundreds of thousands of dollars. All of these factors limit the diversity of environments and targets that can be studied with a MoCap system. We aim to develop a low-cost solution for imitating diverse, in-the-wild motion skills from easily-accessible videos.

Alternatively, markerless motion capture has become increasingly popular in recent years~\cite{
Joo_2017_TPAMI, 
% peng2021neural, 
yoon2021humbi, bala2020openmonkeystudio}. For example, \cite{Joo_2017_TPAMI} built a multi-view video-capture system to capture and reconstruct human motion. Although those systems capture whole-body movement without using markers, they still require an indoor studio with hundreds of synchronized cameras, making it challenging to generalize to in-the-wild targets and behaviors. Some recent works~\cite{kocabas2020vibe, kanazawa2019learning
% , xiang2019monocular
} learn data-driven models to predict full body movements from a single monocular camera. However, they heavily rely on carefully-constructed template models (e.g., SMPL~\cite{loper2015smpl}) and do not generalize well to in-the-wild videos or non-human subjects.

Recent advances in differentiable rendering~\cite{tewari2022advances, mildenhall2021nerf} and robust dense point tracking (e.g. optical flow~\cite{
% bruhn2005lucas, sun2018pwc, 
yang2019volumetric, teed2020raft} and DensePose~\cite{guler2018densepose
% ,
% neverova2020continuous, 
% neverova2021discovering
}) have enabled test-time optimization of dense surface structure and motion given real-life videos~\cite{yang2021lasr,
% wu2021dove,
yang_banmo_2022}. Building on these prior algorithms, our work performs key-point trajectory tracking of human and animal motions in 3D from a single, moving camera
% casual monocular 
video without assuming a predefined shape template.
% \todo{emphasize we reconstructe in world frame from single, moving camera}

\subsection{Trajectory Optimization through Contact}
Trajectory optimization is a powerful tool for designing dynamic behaviors for robotic systems.
% \cite{mellinger2011minimum, kuindersma2016optimization, katz2019mini} 
Given an initial guess, trajectory optimization formulates a nonlinear program (NLP) to solve for an optimal control sequence under robot dynamics and environmental constraints.
% \cite{kelly2017introduction}
This technique has been a major component of important breakthroughs in legged autonomy in recent years \cite{kuindersma2016optimization,bledt_mit_2018,katz2019mini,boston_dynamics_more_2019}.

One of the hardest problems in planning and control for legged robots is reasoning about contact forces and timing as feet make and break contact with the environment, producing discontinuous impact events. A common approach for modeling rigid-body contact interactions is to formulate the dynamics as a linear complementarity problem (LCP) \cite{stewart_implicit_1996, howell_dojo_2022}, which solves for the next system state under impact and friction constraints.
%Intuitively, the normal force constraint enforces that contact forces are only active when the contact distance between two objects is zero. Likewise, the friction cone constraint governs the sticking and sliding modes of frictional contact.
These LCP dynamics can be enforced as constraints in trajectory optimization methods \cite{posa_direct_2014, manchester2019contact}.

If the contact schedule is predefined \cite{bledt_mit_2018}, the dynamics can be written as a hybrid system with known transition times. This method, commonly referred to as hybrid trajectory optimization \cite{
pardo_hybrid_2017, crocoddyl20icra
% ,kong_hybrid_2022
}, can be solved quickly and is effective for periodic gaits on flat terrain. This approach has also enabled diverse locomotion behaviors on robot hardware through motion-template libraries that can then be combined and tracked online to form long-horizon locomotion behaviors over challenging terrains \cite{boston_dynamics_more_2019}.

In contrast, if the contact schedule cannot be determined a priori, the contact interactions must be solved \emph{implicitly}, resulting in a much larger and more challenging nonlinear optimization problem \cite{howell_calipso_2022, noauthor_ipopt_nodate}. This method is referred to as contact-implicit trajectory optimization \cite{posa_direct_2014, 
% patel_contact-implicit_2019, 
manchester2019contact}. Reliably solving contact-implicit trajectory optimization problems to high accuracy is challenging even in offline settings, and the solution can be sensitive to model mismatch. Several previous studies have aimed to improve numerical accuracy \cite{
% patel_contact-implicit_2019, 
manchester2019contact}, convergence properties \cite{howell_calipso_2022}, or robustness to contact model uncertainty \cite{drnach_robust_2021, drnach_mediating_2022}.

% However, this problem is extremely nonlinear and non-convex, rendering real-time solutions not possible. 

In this work, we reason about contact interactions in each stage: During 3D reconstruction, we roll out reconstructed motions in a differentiable simulator \cite{warp2022}, where contact is approximated with a spring-damper model.
% \cite{marhefka1996simulation}. 
In trajectory optimization, we enforce rigid-body contact dynamics using LCP constraints \cite{manchester2019contact}. Finally, during online tracking, we solve a simplified contact-implicit problem that can fully reason about contact mode timing.

\subsection{Online Control for Legged Robots}
Online feedback control has been widely studied for both quadrupedal \cite{bledt_mit_2018} and bipedal \cite{powell_model_2015} robots in recent years. MPC \cite{bledt_mit_2018, cleach_fast_2021} and RL \cite{fu_deep_nodate, smith_walk_2022} have emerged as popular approaches for executing dynamic locomotion behaviors on legged systems.

Different from offline trajectory optimization, MPC typically uses a simplified model and only solves a finite-horizon variant of the optimal control problem to achieve real-time performance. Notably, \cite{bledt_mit_2018} uses a heuristic foothold location scheduler and solves a convex quadratic program (QP) for desired stance-foot forces. This method achieves robust locomotion without any offline computation. A lower-level whole-body controller (WBC) \cite{kim_highly_2019} or inverse kinematics (IK) is then used to map the MPC solution into the robot joint space. At the motor level, a hand-tuned proportional-derivative (PD) controller is used to track this joint-level trajectory on the robot. The offline trajectory optimization and online MPC pipeline have enabled several impressive real-world robot behaviors, such as quadruped jumping \cite{nguyen_continuous_2022, 
% chignoli_online_2021, 
zhou_momentum-aware_2022}, humanoid parkour \cite{boston_dynamics_more_2019}, and even coordinated heterogeneous muti-robot dance routines \cite{boston_dynamics_no_2022}.

% A recent approach\cite{cleach_fast_2021} utilizes the two stages of the optimal control strategy discussed in Section \ref{sec:related-planning-control} and differentiable MPC (cite Zico). In this approach, the first stage trajectory optimization problem contains LCP contact constraints, the solution of which contains not only contact force trajectories but also the derivatives of LCP constraints along the solved state trajectories. In the second stage, the MPC solves a simpler optimization problem where the derivatives of LCP constraints are used to adjust contact forces if the robot state deviates from the reference. 

Model-free RL aims to learn a feedback control policy, typically represented by a deep neural network, by collecting experience in a simulated environment or the real world. The learned policy is optimized by maximizing a reward function using gradient descent algorithms. Similar to model-based pipelines, recent RL methods also rely on hand-tuned low-level PD controllers to execute behaviors in the real world \cite{fu_deep_nodate, smith_walk_2022}.
RL has also been successfully applied to animal imitation from motion capture reference data on a quadrupedal robot\cite{bin_peng_learning_2020}.

In this work, we adopt a general-purpose contact-implicit MPC (CI-MPC) algorithm \cite{cleach_fast_2021}, capable of controlling both quadruped and humanoid robots, and pair it with a low-level joint-space controller based on IK and PD feedback for hardware execution. Our contact-implicit MPC formulation can reason about contact timing and forces in real time, enabling robust tracking of complex behaviors.
% \vspace{-1mm}
\begin{table}
    \centering
    \caption{A comparison between recent motion imitation methods for legged systems}

    \begin{tabular}{c c c c c c|c}
        \toprule
         &  \cite{peng_deepmimic_2018} & \cite{bin_peng_learning_2020} &  \cite{li_fastmimic_2021} & \cite{yao_imitation_2022} &\cite{
         % kang_animal_nodate, 
         kang_animal_2021}  & SLoMo\\
         \toprule 
         Single RGB Camera            & \xmark & \xmark & \xmark & \cmark & \xmark & \cmark\\
         % \toprule
         No Manual Label       & \cmark & \cmark & \xmark & \cmark & \cmark & \cmark\\
         % Explainable Planner   & \xmark & \xmark & \xmark & \cmark & \cmark & \cmark\\
         Robot Hardware        & \xmark & \cmark & \cmark & \cmark & \cmark & \cmark\\
         System Agnostic       & \cmark & \xmark & \xmark & \xmark & \xmark & \cmark\\
         % Uneven Terrain        & \cmark & \xmark & \xmark & \xmark & \xmark & \cmark \\
         \toprule
    \end{tabular}
    \label{tab:imitation_comp}
    \vspace{-5mm}
\end{table}

\subsection{Motion Imitation}
Imitating motion primitives from nature can be an effective strategy for producing natural-looking robot behaviors while avoiding tedious, manual trajectory design. Researchers have studied various approaches for mimicking human motions \cite{pollard2002adapting, koenemann2014real}. For example, learning-from-observation (LFO) converts MoCap motion sequences into reference motion primitives that can be tracked by a zero-moment point (ZMP) controller, which enables a humanoid robot to graciously perform traditional Japanese dance routines \cite{nakaoka2007learning}. 

Recently, imitation learning has been used to produce animal-like movements for animation \cite{peng_deepmimic_2018}. \cite{bin_peng_learning_2020, kim_human_2022, yao_imitation_2022} demonstrated imitated motions where the sim-to-real gap was bridged through online adaptation on a quadrupedal robot. Similarly, model-based imitation \cite{kang_animal_2021, li_fastmimic_2021} has also produced animal-like locomotion on a real-world quadrupedal robot using hybrid trajectory optimization, where foothold locations and timing were predetermined by thresholding MoCap data. These prior works rely on marker-based motion capture to acquire motion priors. \cite{yao_imitation_2022} uses RGB videos but still relies on manual labeling to acquire trajectories. As discussed in Section \ref{sec:motion-capture}, these methods have some major limitations. Recent videos from Boston Dynamics \cite{boston_dynamics_no_2022} demonstrate impressive robot dancing through complicated choreography design, character animation, and robot trajectory optimization procedures; an expensive, time-consuming process. 

In robot manipulation, recent work \cite{bahl_human--robot_2022, sivakumar_robotic_2022} takes essential steps towards acquiring in-the-wild robotic skills from real-world RGB videos through reinforcement learning. The key insight from these two studies is that computer-vision techniques can now process widely-available human and animal footage into representations that can be very effective priors for acquiring robotic skills. In this work, we tackle the problem of imitating locomotion skills by extracting target motions from videos using 3D reconstruction. Different from previous RL-based imitation methods that only consider robot kinematics, we use optimal control methods that allow us to explicitly reason about dynamics. 

% We propose a simpler and more cost-effective alternative: generating feasible robot trajectories directly from widely available ``casual'' monocular videos. A qualitative comparison of SLoMo to previous work on motion imitation is presented in Table \ref{tab:imitation_comp}.
We propose a simple and cost-effective method: synthesize legged robot motion primitives directly from casual RGB videos without manual labels and leverage a model-based controller for robust execution on robot hardware. Additionally, we demonstrate that our control strategy generalizes across quadruped and humanoid robots. Moreover, the model-based approach allows us to interpret the output trajectories and explicitly reason about hardware limitations like torque limits.
\section{Video-to-Robot Motion Transfer}\label{motion_transfer}

In this section, we present the SLoMo algorithm for robot motion imitation from in-the-wild videos: Section \ref{method:reconstruction} (Fig. \ref{fig:method_overview} (a)) describes the 3D reconstruction pipeline for generating physically-plausible key-point trajectories from casual footage. Section \ref{method:contact-implicit_traj_opt} (Fig. \ref{fig:method_overview} (b)) explains the offline trajectory-optimization problem used for constructing dynamically feasible robot reference trajectories and foot-contact sequences that mimic key-point movements. Section \ref{method:ci-mpc} (Fig. \ref{fig:method_overview} (c)) details the contact-implicit MPC algorithm\cite{cleach_fast_2021} for tracking the optimal reference online. 

\subsection{Physics-Informed Reconstruction from Casual Videos}\label{method:reconstruction}

Given videos of a target animal or human, our goal is to estimate its kinematic key-point trajectory in the world coordinates. Similar to prior work \cite{yang_banmo_2022, yang2021lasr, yang2023ppr}, we simultaneously reconstruct articulated shapes and kinematic skeleton trajectories, connected by a blend skinning model. To reconstruct physically plausible trajectories, we set up a physics-informed optimization by coupling differentiable rendering costs with an additional physics-roll-out cost.
% For the remainder of this section, 

We define the ${\bf x}(t)\in\mathbb{R}^{3N}$ to be the estimated key-point trajectory, where $N$ is the total number of key-points, $t\in\{1\dots T\}$ to be discrete time steps, and $T$ to be the number of frames in a given video.

\textbf{Shape Model:} 
We use a Multi-Layer Perceptron (MLP) parameterized by $\sigma$ to represent the visual properties of the default state of the object:
\begin{align}
    (d,{\bf c}) &= \mathbf{MLP}_\mathbf{\sigma}({\bf X}), \label{eq:mlp_shape}
\end{align}
where $d \in \mathbb{R}$ is the assigned signed distance and $\mathbf{c} \in \mathbb{R}^3$ is the color at each point $X \in \mathbb{R}^3$. The points that have distance $d=0$ are on the surface of the object. This representation is similar to a neural radiance field (NeRF) \cite{mildenhall2021nerf} except we remove the view dependence of color.

\begin{figure}
    \centering
    \includegraphics[width=\linewidth]{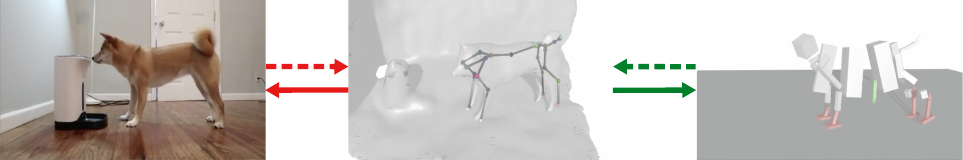}
    \caption{The physics-informed 3D reconstruction pipeline: Differentiable rendering from a monocular video (left) and differentiable physics simulation (right) update key-point motion estimates (middle). Solid red and green arrows represent rendering and physics-roll-out costs; dashed arrows are corresponding gradients.}
    \label{fig:reconstruction}
    \vspace*{-7mm}
\end{figure}

% \noindent{\bf Key-point Skeleton.} 
\textbf{Kinematic Skeleton Model}: To model the motion of the subject animal or human, we use a predefined target kinematic skeleton model consisting of $B$ rigid links and $N$ spherical joints, among which a root link is selected to define the model's location and orientation in space. For simplicity, we utilize joint locations as key points (Fig. \ref{fig:reconstruction}).
The key-point trajectory ${\bf x(t)}$ can be calculated using forward kinematics. When the skeleton changes configuration, the object's shape should deform and move along with the skeleton. This motion is described by a neural blend-skinning model $\mathcal{W}_t(\bf X)$ \cite{yang_banmo_2022} as an MLP parameterized by $\lambda$:
\begin{align}
    \mathcal{W}_t(\bf X) &= \mathbf{MLP}_\mathbf{\lambda}({\bf X}; Q, G, t). \label{eq:mlp_transformation}
\end{align}
The neural blend-skinning model warps 3D points of the new configuration to the default configuration, which can then be used to query the visual properties of the object.
We further parameterize the joint angles $Q=\mathbf{MLP}_\kappa(t)$ and SE(3) root-link transformation $G=\mathbf{MLP}_\eta(t)$. Then, after training, the zero-level set of $d$ given by  
\begin{align}
(d, {\bf c}) = \mathbf{MLP}_\mathbf{\sigma}(\mathcal{W}_t(\bf X))   \label{eq:mlp_final} 
\end{align}
% represents the surface of the object in the time $t$ configuration.
represents the surface of the object at time $t$.
% \noindent{\bf Differentiable Volume Rendering.} 

\textbf{Differentiable Volume Rendering}: In addition to the object shape model, we train another parameterized background scene model similar to Eq.~\eqref{eq:mlp_shape}. The shape and scene models can then be used together to differentiably render images given a camera's view transformation and intrinsic parameters~\cite{niemeyer2021giraffe}. We achieve this by performing ray casting for each image pixel $p$. For all 3D points along the ray within a given distance range, we use Eq.~\eqref{eq:mlp_final} to query their 3D color and density and perform a weighted average to compute $\hat{{\bf c}}_t(p)$, the pixel values on the rendered image, including 2D color and object silhouette. We further render optical flow from $t$ to $t+1$ by warping and projecting the queried points from the current configuration to the next frame configuration.

% \noindent{\bf Rendering Cost.} 
\textbf{Rendering Cost}: We compare the expected color, object silhouette, and optical flow of pixels $\hat{{\bf c}}_t(p)$ on the rendered image with the input video observations at time $t$. The observation $\bar{{\bf c}}_t(p)$ comes from basic image-processing and segmentation methods \cite{kirillov2020pointrend, yang2019volumetric}. A rendering cost function is defined as:
\begin{equation}
    \mathcal{L}_{render} = \sum_{t}\sum_{p}\| \hat{{\bf c}_t}(p) -  {\bar{\bf c}_t}(p) \|^2
\end{equation}
We use its gradients to update the model parameters $\sigma$, $\lambda$, $\kappa$, and $\eta$ with the Adam optimizer \cite{kingma2014adam}. 

\textbf{Physics Roll-Out Cost}: Using only the rendering cost, the trained shape model can generate motions from the same view point~\cite{yang_banmo_2022}. However, these motions are often physically unrealistic due to piece-wise scale ambiguity at each individual patch~\cite{
% hartley2003multiple, 
kumar2019superpixel} --- a common issue for reconstructing monocular videos. For example, a dog can be up close and floating in the air or far away and on the ground. Both are equally valid from the same monocular visual evidence, but only the latter obeys physics. To resolve this ambiguity, we introduce a physics-roll-out cost in the optimization problem to encourage a physically plausible rendering solution.

Treating the key-point skeleton as a floating-base multi-rigid-body system, we denote its generalized coordinates as $q(t)$
% \cite{mistry2010inverse}
. During differentiable rendering, using the latest parameters $\kappa$ and $\eta$, we can generate a reference trajectory $q_d(t)$ with $t \in \{1\dots T\}$. Then, in a differentiable simulator \cite{warp2022}, we simulate the multi-body key-point skeleton model with a simple PD controller attempting to track $q_d(t)$ to compute the following cost,
\vspace{-1mm}
\begin{equation}
    \mathcal{L}_{physics} = \sum_{t}\| q(t) - q_d(t) \|^2.
    \vspace{-1mm}
\end{equation}
This cost function is differentiable with respect to $\kappa$, $\eta$, and $\xi$. Note that, in this stage, the physics cost only encourages physical realism in a ``soft'' way and does not guarantee dynamic feasibility like later stages in SLoMo. Thus, we deem the rendered key-point trajectory physically \emph{plausible}. 

\textbf{Coordinate-Descent Optimization}: In theory, the cost $\mathcal{L}_{render}+\mathcal{L}_{physics}$ can be optimized together. However, in practice, the volume rendering and the physics simulator run at different frequencies, making joint optimization inefficient. Instead, we use coordinate descent to alternately minimize each cost function. In each iteration, we first minimize the rendering cost, during which we regularize $Q$ and $G$'s output toward the previous simulator-generated trajectory. We then perform a physics roll-out optimization step, where only physics parameters $\xi$ are updated. We also use other strategies to improve convergence such as over-parameterization
% \cite{arora2018optimization}% and adding a cycle consistency cost~\cite{li2021neural}
. At convergence, we take the rendered key-point trajectories $x^*(t)$ as output to the next stage of the motion-imitation pipeline. More details of the reconstruction optimization can be found in \cite{yang_banmo_2022, yang2023ppr}. 
% \vspace{-6mm}
\subsection{Contact-Implicit Trajectory Optimization}\label{method:contact-implicit_traj_opt}
% \vspace{-2mm}
After generating key-point trajectories from monocular videos, we solve a trajectory-optimization problem subject to robot dynamics and contact constraints. In particular, we use contact-implicit trajectory optimization \cite{posa_direct_2014, manchester2019contact}, which jointly solves for robot states, controls, and contact forces with a direct collocation formulation. Compared to hybrid trajectory optimization \cite{pardo_hybrid_2017, crocoddyl20icra}, the contact-implicit method does not require a predefined contact sequence, which is an important advantage in our motion-imitation workflow since the reconstructed key-point trajectories can suffer from dynamically infeasible artifacts (e.g. foot sliding), even with the physics roll-out cost, which makes applying a heuristic contact schedule impractical. By using the contact-implicit formulation, we allow the optimizer to automatically generate a feasible robot gait sequence and corresponding reference trajectory that is similar to the key-point trajectory without separately predefining a contact schedule. 

The offline contact-implicit trajectory-optimization problem has the following form,
\begin{mini}|l|
  {\mathcal{H}, \mathcal{X}, \mathcal{U}, \mathcal{\lambda}}{\sum_{t=1}^{T-1} \frac{h_t}{2} \Big[ (x_t - x_t^*)^\top Q (x_t-x_t^*) + u_t^\top R u_t \Big] }{}{}
  \breakObjective{ + (x_N - x_N^*)^\top Q_N (x_N-x_N^*)} 
  \addConstraint{x_{t+1}}{= \textbf{NCP}_t(h_t,x_t, u_t, \lambda_t)}
  \addConstraint{u_t}{\leq u_{max}} 
  \addConstraint{u_t}{\geq u_{min},}
  \label{eq:optimal_control}
\end{mini}
where $h_t$ is the time step, $x_t = (q_t, v_t)$ is the state, and $u_t$ are the controls, and $\lambda_t$ are contact forces at time $t$, respectively. $\textbf{NCP}(x, u)$ represents the robot's nonlinear dynamics, including contact constraints, as a nonlinear complementarity problem \cite{manchester2019contact,howell_dojo_2022}. We optimize a quadratic tracking cost where $x_t^*$ is the reconstructed key-point state at time $t$ and $Q$ and $R$ are diagonal weighting matrices. This formulation allows the solver to infer a contact sequence and corresponding robot trajectories by imitating noisy and dynamically infeasible key-point data. We refer to a solution of \eqref{eq:optimal_control} as a \emph{reference trajectory}.
% \vspace*{-0mm}
\subsection{Contact-Implicit Model-Predictive Control}\label{method:ci-mpc}
% \vspace*{-5mm}
To stabilize and track reference trajectories in real-time, we use contact-implicit model-predictive control \cite{cleach_fast_2021}. CI-MPC is a general method for controlling robots that make and break contact with their environment. Different from standard convex MPC approaches, CI-MPC models the robot's dynamics with a time-varying linear complementarity problem. This LCP can be thought of as a local approximation of the nonlinear complementarity problem in \eqref{eq:optimal_control} about the reference trajectory, which makes it computationally easier to solve. Importantly, however, it maintains the ability to reason about discontinuous contact-switching events.

The CI-MPC tracking problem is:
% \begin{mini}|l|
%   {\mathcal{X}, \mathcal{U}}{\sum_{t=1}^{H} \frac{1}{2} \Big[ (x_t - \Bar{x}_t)^\top Q (x_t-\Bar{x}_t) \Big_ }{}{} \label{eq:model-predictive-control}
%   \breakObjective{\Big_ + (u_t - \Bar{u}_t)^\top R (u_t - \Bar{u}_t) \Big]}
%   \addConstraint{x_{t+1}}{= \textbf{LCP}_t(x_t, u_t)}
%   \addConstraint{u_t}{\leq u_{max}} 
%   \addConstraint{u_t}{\geq u_{min},}
% \end{mini}
\begin{mini}|l|
    {\mathcal{X}, \mathcal{U}}{\sum_{t=1}^{H} \frac{1}{2} \Big[ (x_t - \Bar{x}_t)^\top Q (x_t-\Bar{x}_t) }{}{} \label{eq:model-predictive-control}
  \breakObjective{+ (u_t - \Bar{u}_t)^\top R (u_t - \Bar{u}_t) \Big]}
  \addConstraint{x_{t+1}}{= \textbf{LCP}_t(x_t, u_t)}
  \addConstraint{u_t}{\leq u_{max}} 
  \addConstraint{u_t}{\geq u_{min},}
\end{mini}
where $x_t$, $u_t$ are state and control decision variables and $\Bar{x}_t$, $\Bar{u}_t$ are the reference state and control trajectories from \eqref{eq:optimal_control} and $H$ is the MPC horizon, which is generally shorter than the full length $T$ of the reference trajectory to enable faster online solution times. The interested reader is referred to \cite{cleach_fast_2021} for more details on CI-MPC.

% \begin{figure*}[h]
%     \centering
%     \hspace{0.1cm}
%     \includegraphics[height=5.0cm]{figures/dog_reach.pdf}
%     \hfill
%     \includegraphics[height=5.0cm]{figures/cat_walk_freeze_rame.pdf}
%     \hspace{0.1cm}
%     \caption{Freeze-frame comparisons between input video (top row), reconstructed key points (second row), trajectory optimized reference (third row), and online tracking on the Unitree Go1 robot (bottom row) for \emph{dog reach} (left) and \emph{cat pace} (right).}
%     \label{fig:reach_cat_walk_freeze_frame}
% \end{figure*}

\section{Experiments and Results}\label{results}

In this section, we present the results of experiments demonstrating the capabilities of SLoMo on a variety of robotic systems in simulation and on hardware. In particular, we demonstrate two important features of our method: 1) our entire framework can successfully transfer motions from casual videos to robot hardware and 2) our pipeline is model agnostic and can support both quadruped and humanoid robots. We carefully identify canonical legged robot movements that require no contact switching, periodic contact switching, and dynamic non-periodic contact switching.
Experiment videos are available on our \href{https://slomo-www.github.io/website/}{website}. 
An implementation of the framework can be found on \href{https://github.com/johnzhang3/SLoMo}{GitHub}.
% \begin{center}
    
% \end{center}
\vspace{-2mm}
\subsection{Experimental Setup}
Our 3D reconstruction stage is implemented with PyTorch. Processing a one-minute video takes eight hours on a computer with 8 NVIDIA GeForce RTX 3080 GPUs. We run offline trajectory optimization and online MPC on a workstation computer equipped with an Intel i9-12900KS CPU and 64GB of memory. This workstation computer is connected to the robot via Ethernet. All hardware experiments are run on a Unitree Go1 quadruped robot.
% \vspace{-3mm}
\begin{figure}
    \centering
    % \includegraphics[width=\textwidth]{figures/freeze_frame.pdf}
    % \hspace{0.1cm}
    \includegraphics[height=3.3cm]{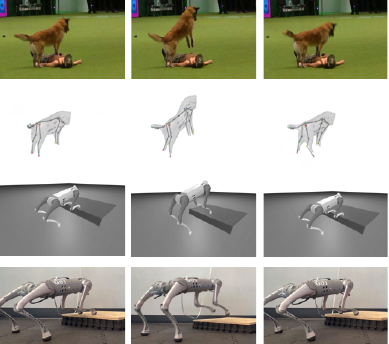}
        \hfill% \hspace{}
    \includegraphics[height=3.3cm]{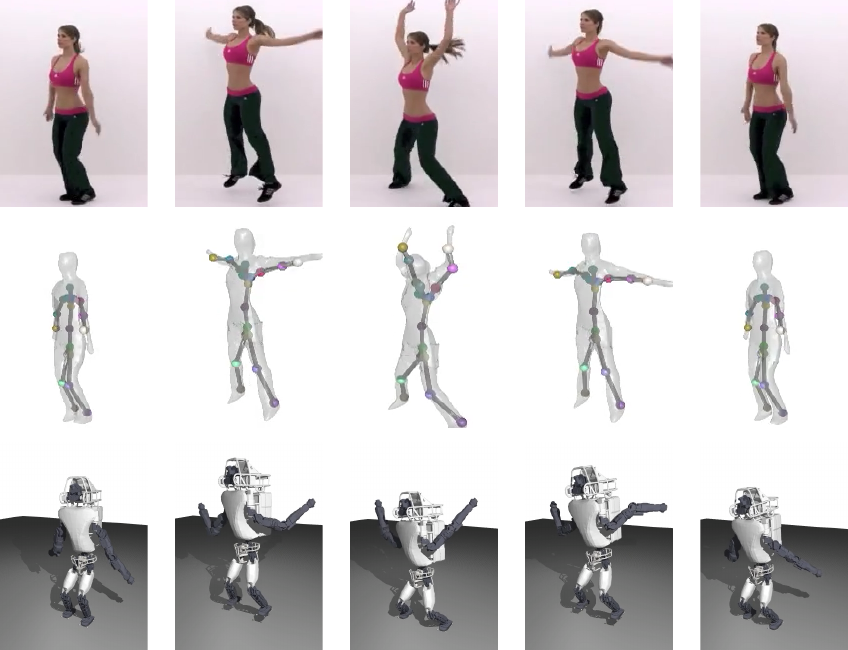}
    % \hspace{0.3cm}
    % \hfill
    % \includegraphics[height=5.0cm]{figures/human_wave.pdf}
    % % \hspace{0.1cm}
    \caption{\emph{Dog CPR} (left), \emph{Human Jumping Jack} (right) motion imitation demonstrations. 
    The top row shows frames from the original video, the second row shows the key-point trajectory, and the third row shows the dynamically feasible trajectory being executed by the Unitree Go1 and Atlas robots. The bottom row on the left shows \emph{dog CPR} on hardware.
    % under terrain height mismatch.
    }
    \label{fig:cpr_jumping_freeze_frame}
    \vspace{-5mm}
\end{figure}
\subsection{Quadruped}

We verify our video-to-robot motion transfer approach on three video input examples: a dog reaching for water (\emph{dog-reach}) --- shown in Fig. \ref{fig:teaser-demo} (left); a cat pacing across a living room (\emph{cat pace}) --- shown in Fig. \ref{fig:teaser-demo} (second to left); and a dog performing CPR on a human (\emph{dog CPR}) --- shown in Fig. \ref{fig:teaser-demo} (second to right). 
% Additionally, we demonstrate controller robustness to online terrain height model mismatch while performing a dynamic push-up motion in the \emph{dog CPR} example. 
% To the best of our knowledge, prior work \cite{bin_peng_learning_2020, kang_animal_2021, kang_animal_nodate} only imitates animal motions on a flat ground; the proposed framework is the first motion-imitation method to show robustness to terrain mismatch while imitating dynamic animal behaviors. 

\textbf{Point-foot quadruped model}: We use a simplified robot dynamics model for offline trajectory optimization and online control. This point-foot representation of the robot dynamics neglects the leg dynamics and instead models the feet as point masses. The model has 18 degrees of freedom and 12 control inputs. The controls are modeled as three-dimensional internal forces between each foot and the body. Note that this model is different from the skeleton model used during reconstruction, which contains additional links and joints (e.g. legs, tails, torso, etc.). 

\textbf{Hardware setup}: For the hardware experiments, a hand-tuned PD controller running at $1000$ Hz is used to track the forces and foot positions computed by the MPC policy on the Unitree Go1 robot. State estimation is also provided at $1000$ Hz with a Kalman Filter that utilizes robot joint encoders and an onboard IMU. We take time discretization of $0.05$ s offline and track the reference with online MPC at $100$ Hz with a prediction horizon of $0.15$ s on the Unitree Go1 robot. 

\textbf{Dog reach}: We take a video clip of a dog reaching for an automatic water feeder and compute an offline reference trajectory that is $5.0$ s long. (Fig. \ref{fig:teaser-demo} left).

\textbf{Cat pace}: We take a video clip of a cat pacing across a living room (Fig. \ref{fig:teaser-demo} second to left) and compute an offline reference trajectory that is $3.0$ s long, then repeated to form a continuous forward walking gait.
% We track this reference on the Unitree Go1 robot with MPC horizon $0.10$ seconds and update frequency $100$ Hz (Fig. \ref{fig:reach_cat_walk_freeze_frame} right). 

\begin{figure}[h]
    \centering
    \setlength\fwidth{0.50\textwidth}
    \setlength\fheight{0.40\textwidth}
    \resizebox{\linewidth}{!}{\input{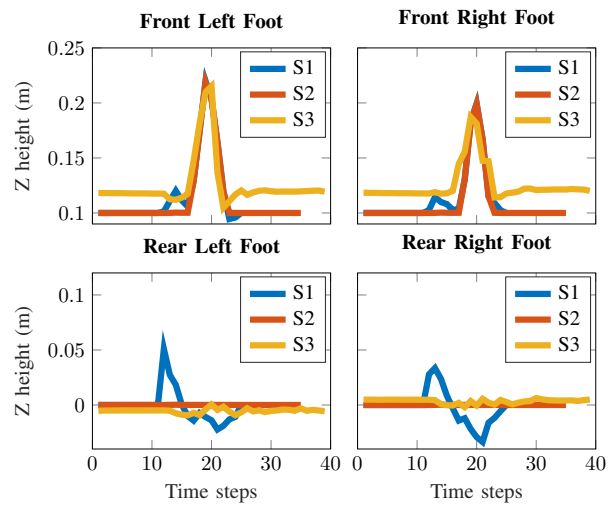}}
    % \vspace{-9mm}
    \caption{Foot height trajectories comparison for the dog-CPR experiment. Each plot corresponds to one foot of the robot. The blue lines are trajectories generated by the 3D reconstruction stage (S1). The red lines are the outputs of the trajectory optimization stage (S2). The yellow lines are the state feedback on robot hardware showing the result foot trajectories of the MPC policy (S3). }
    \label{fig:foot_height}
    \vspace{-3mm}
\end{figure}

\textbf{Dog CPR}: We take a video clip of a dog performing CPR on a human (Fig. \ref{fig:cpr_jumping_freeze_frame} left) and compute an offline reference trajectory that is $0.65$ s long 
% with a time discretization of $0.05$ seconds. We track this reference on the Unitree Go1 robot with MPC horizon $0.15$ seconds and update frequency $100$ Hz. 
We model the terrain as a step where the front feet plan for a contact distance $0.1$ m higher than the back feet. This motion primitive is fairly dynamic and would be difficult to design manually. Fig. \ref{fig:foot_height} compares the output trajectories of each stage of SLoMo for each foot, where the key-point trajectory (blue) contains infeasible ground penetration, while the optimized reference (red) eliminates this unphysical artifact. 
% Although the nominal front feet height is 0.1 m in the reference, we use a 0.12 m tall step during the hardware experiment.
% The MPC policy adapts online for contact height mismatch and demonstrates robust performance, as shown in the supplementary video.

\subsection{Humanoid}
We show that our framework can also be applied to imitating human movements on humanoid robots in simulation on the Atlas robot.

\textbf{Point-foot humanoid model}: We use a simplified model to represent the dynamics of the humanoid robot. We model each foot as a rectangular prism with two contact points: one at the toe and the other at the heel. We similarly model each hand as a point mass. The model has 24 degrees of freedom and 18 control inputs. We use time discretization of $0.1$ s for the experiments below.

% \emph{Human-waving} In this example, we take a video of human waving at the camera. 

\textbf{Human Stretching}: We take a video of a human raising both arms in a stretching motion (Fig. \ref{fig:teaser-demo} second to left) and compute an offline reference trajectory that is $3.8$ s long.

% \textbf{Human Waving}: We take a video of a human waving at the camera (Fig \ref{fig:cpr_stretch_wave_freeze_frame} right) and compute an offline reference trajectory that is $4.0$ s long.

\textbf{Human Jumping Jacks} We take a video of a human performing a jumping jack exercise (Fig. \ref{fig:cpr_jumping_freeze_frame} right) and compute an offline reference trajectory that is $1.5$ s long.
\vspace{-5mm}

\subsection{Comparisons to an RL Policy}
\begin{figure}
    \centering
    \includegraphics[trim=1cm 1cm 1.0cm 3.0cm, clip, width=\linewidth]{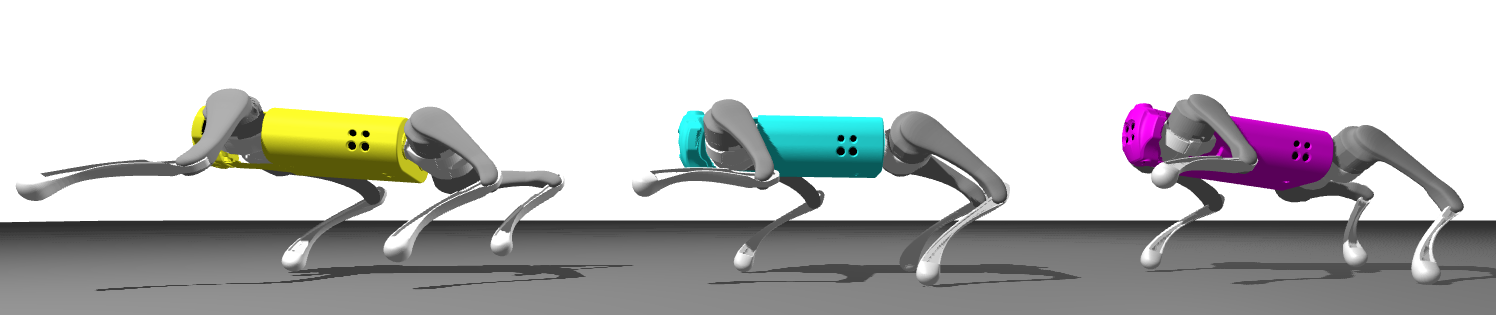} \\ 

    \caption{A comparison between the reconstructed key-point reference (yellow), optimized reference (teal, ours), and learned imitation policy \cite{bin_peng_learning_2020} (purple).}
    \label{fig:RL_comp}
    \vspace{-5mm}
\end{figure}

We show that it is also possible to replace the trajectory-optimization and MPC steps of our method with an RL method \cite{bin_peng_learning_2020} in simulation. We train RL policies for the Dog Reach and Cat Pace examples with $4$ random seeds each. In Fig. \ref{fig:RL_comp}, we showcase the highest-performing policy on the robot. Notably, while the best policy shows reasonable performance, we find that the RL policies exhibit a high degree of variance across different seeds in the Dog Reach example. This high variance makes fair, rigorous comparisons between model-based optimization and model-free RL difficult. However, we still believe imitating animal movements using RL is a promising approach and deserves further investigation.

% \vspace{-5mm}

% \subsection{Monocular video motion transfer using RL}
% We show that our proposed video-to-robot motion transfer can be generalized to RL-based imitation learning algorithms. Different from \cite{bin_peng_learning_2020} which uses clean mocap data, we show that we can also use key-point trajectories from Section \ref{sec:motion-capture} for imitation learning. \todo{refer to figures and potentially training curves}
% \vspace{-2mm}
\section{Discussion and Conclusions}\label{conclusions}
% We present SLoMo, a first-of-its-kind framework for enabling legged robots to imitate animal and human motions captured from real-world casual monocular videos. We found that 3D reconstruction \cite{yang_banmo_2022, yang2023ppr} can be a surprisingly powerful tool for acquiring physically-plausible motion trajectories from just RGB videos. While these trajectories are still noisy and dynamically infeasible, a trajectory optimizer \cite{manchester2019contact} or RL \cite{bin_peng_learning_2020} can be applied to plan for dynamically feasible trajectories that can be safely executed on a given robot. We use a contact-implicit trajectory optimization method to reason about contact events and timing, and a contact-implicit MPC that is robot-agnostic and handles non-periodic contacts. This specific offline trajectory optimization and MPC combination facilitate motion transfer regardless of behavior periodicity. 

We present SLoMo, a first-of-its-kind framework for enabling legged robots to imitate animal and human motions captured from real-world casual monocular videos. Our research highlights that recent advancements in 3D reconstruction \cite{yang_banmo_2022, yang2023ppr} are effective at extracting physically plausible motion trajectories solely from monocular RGB videos. These trajectories are quite noisy and dynamically infeasible, even with the physics-based roll-out cost, making them unsafe to execute directly on real robots. To overcome this, an off-the-shelf trajectory optimizer \cite{manchester2019contact} or RL method \cite{bin_peng_learning_2020} is necessary to plan for dynamically feasible trajectories that track the retrieved motions from videos. 
% In fact, we show that the resulting trajectories from trajectory optimization can be safely executed on a given robot with an MPC controller. However, note that the offline trajectory optimizer/RL was a necessary step. Directly tracking the noisy/dynamically infeasible trajectories using an MPC controller led to very unstable/unsafe behaviors. 

In this paper, we use a contact-implicit trajectory optimization method to reason about contact events and timing, and an MPC that is robot-agnostic and handles non-periodic contacts. This specific offline trajectory optimization and MPC combination facilitates motion transfer regardless of behavior periodicity, % as opposed to more standard MPC approaches such as convex-MPC which only works with periodic gaits. 
allowing us to generalize to arbitrary human and animal behaviors.

\subsection{Limitations}
SLoMo is a promising first step towards imitating human and animal behaviors on real-world robots from in-the-wild video footage. However, several limitations remain that should be addressed in future research: First, we make key model simplifications and assumptions in Sections \ref{method:contact-implicit_traj_opt} and \ref{method:ci-mpc} by using a point-foot model to represent the quadruped and humanoid dynamics. It should be possible to extend this work to use full-body dynamics in both offline and online optimization steps to fully leverage a robot's capabilities. For example, humans and animals are capable of making contact with the world in very rich ways (e.g. a dog can use its head to move an object), but executing such behaviors for legged robots remains an open research problem. Second, the reconstruction step is computationally expensive, and we manually scale the reconstructed character to better match the kinematics of the target robot in Section \ref{method:reconstruction}. It should be possible to automate this scaling in Problem \ref{eq:optimal_control}. Addressing morphological differences between video characters and corresponding robots in a principled, automated manner and accelerating reconstruction will be crucial for scaling our framework to large video datasets.
 
\subsection{Future Work}
Many exciting directions for future research remain: First, several trade-offs should be investigated in which components of our framework are swapped out. For example, leveraging RGB-D video data can likely improve reconstruction quality at the expense of data availability.
Secondly, it should be possible to deploy the SLoMo pipeline on humanoid hardware, imitate more challenging humanoid behaviors, and execute behaviors on more challenging terrains where humans and animals have demonstrated highly athletic and robust behaviors but manually designing trajectories for robots can be challenging. Finally, this work can benefit from a proven, high-performance online control stack (e.g. whole-body control \cite{kim_highly_2019}). We believe that real-world visual data can be a rich source of robot behaviors, and taking advantage of recent advancements in reasoning about such visual data can be extremely powerful for robotics.
% Finally, a natural extension of this work is to use reconstructed motions from videos as priors for learning control policies with reinforcement or supervised learning. 


%%%%%%%%%%%%%%%%%%%%%%%%%%%%%%%%%%%%%%%%%%%%%%%%%%%%%%%%%%%%%%%%%%%%%%%%%%%%%%%%

%%%%%%%%%%%%%%%%%%%%%%%%%%%%%%%%%%%%%%%%%%%%%%%%%%%%%%%%%%%%%%%%%%%%%%%%%%%%%%%%



\section*{ACKNOWLEDGMENTS}
The authors would like to thank Taylor A. Howell, Simon Le Cleac'h, and members of the Robotic Exploration Lab at CMU for their insightful discussions and feedback.

% \vspace{-3mm}
%%%%%%%%%%%%%%%%%%%%%%%%%%%%%%%%%%%%%%%%%%%%%%%%%%%%%%%%%%%%%%%%%%%%%%%%%%%%%%%%
% \bibliographystyle{IEEEtran}
\printbibliography
% \bibliography{references}
\end{document}